# Supervising Unsupervised Learning

**Vikas K. Garg** (MIT) [1]  **Adam T. Kalai** (Microsoft Research) [2]

## Abstract

We introduce a framework to leverage knowledge acquired from a repository of (heterogeneous) supervised datasets to new unsupervised datasets. Our perspective avoids the subjectivity inherent in unsupervised learning by reducing it to supervised learning, and provides a principled way to evaluate unsupervised algorithms. We demonstrate the versatility of our framework via simple agnostic bounds on unsupervised problems. In the context of clustering, our approach helps choose the number of clusters and the clustering algorithm, remove the outliers, and provably circumvent the Kleinberg's impossibility result. Experimental results across hundreds of problems demonstrate improved performance on unsupervised data with simple algorithms, despite the fact that our problems come from heterogeneous domains. Additionally, our framework lets us leverage deep networks to learn common features from many such small datasets, and perform zero shot learning.

## 1. Introduction

Unsupervised Learning (UL) is an elusive branch of Machine Learning (ML), including problems such as clustering and manifold learning, that seeks to identify structure among unlabeled data. UL is notoriously hard to evaluate and inherently undefinable. To make this point as simply as possible, consider clustering the points on the line in Figure 1 (left). One can easily justify 2, 3, or 4 clusters. As (Kleinberg, 2003) argues, it is impossible to give an axiomatically consistent definition of the "right" clustering. However, now suppose that one can access a bank of prior clustering problems, drawn from the same distribution as the current problem at hand, but for which ground-truth labels are available. In this example, evidence may favor two clusters since the unlabeled data closely resembles two of the three 1-dimensional clustering problems, and all the clusterings share the common property of roughly equal size clusters. Given sufficiently many problems in high dimensions, one can learn to extract features of the data common across problems to improve clustering.

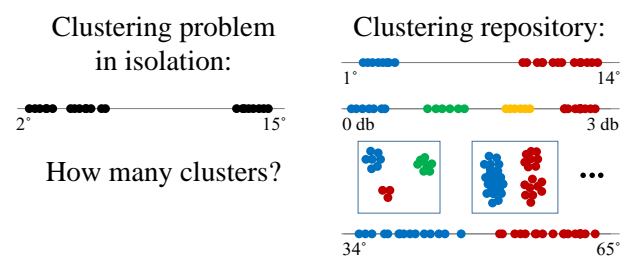

*Figure 1.* In isolation (left), there is no basis to choose a clustering – even points on a line could be clustered in 2-4 clusters. A repository of clustering problems with ground-truth labels (right) can inform the choice amongst clusterings or even offer common features for richer data.

We model UL problems as representative samples from a meta-distribution, and offer a solution using an annotated collection of prior datasets. Specifically, we propose a *meta-unsupervised-learning* (MUL) framework that, by considering a distribution over unsupervised problems, reduces UL to Supervised Learning (SL). Going beyond transfer learning, semi-supervised learning, and domain adaptation (Pan & Yang, 2010; Kingma *et al.*, 2014; Siddharth *et al.*, 2017; Rohrbach *et al.*, 2013; Ganin & Lempitsky, 2015; Long *et al.*, 2016; Bousmalis *et al.*, 2016) where problems have the same dimensionality or at least the same type of data (text, images, etc.), our framework can be used to improve UL performance for problems of different representations and from different domains.

Empirically, we train meta-algorithms on the repository of classification problems from `openml.org` which has a variety of datasets spanning domains such as NLP, computer vision, and bioinformatics, among others. Ignoring labels, each dataset can be viewed as a UL problem. We take a data-driven approach to quantitatively evaluate and design new UL algorithms, and merge classic ideas like unsupervised learning and domain adaptation in a simple way. This enables us to make UL problems, such as clustering, well-defined.

[1]CSAIL, MIT  [2]Microsoft Research. Correspondence to: Vikas K. Garg <vgarg@csail.mit.edu>, Adam T. Kalai <adum@microsoft.com>.

Our model requires a *loss* function measuring the quality of a solution (e.g., a clustering) with respect to some conceptual *ground truth*. Note that we require the ground truth labels for a training repository but not for test data (where they may be impossible to acquire). We suppose that we have a repository of datasets, annotated with ground truth labels, that is drawn from a *meta-distribution* $\mu$ over problems, and that the given data $X$ was drawn from this same distribution (though without labels). From this collection, one could, at a minimum, learn which clustering algorithm works best, or, even better, which type of algorithm works best for which type of data. The same *meta*-approach could help in selecting how many clusters to have, which outliers to remove, and so forth.

Our theoretical model is a meta-application of Agnostic Learning (Kearns *et al.*, 1994) where we treat each entire labeled problem analogous to a training example in a supervised learning task. We show how one can provably learn to perform UL as well as the best algorithm in certain classes of algorithms. Our work also relates to supervised learning questions of meta-learning, sometimes referred to as auto-ml (e.g. Thornton *et al.*, 2013; Feurer *et al.*, 2015), learning to learn (e.g. Thrun & Pratt, 2012), Bayesian optimization (e.g. Snoek *et al.*, 2012) and lifelong learning (e.g. Thrun & Mitchell, 1995; Balcan *et al.*, 2015). In the case of supervised learning, where accuracy is easy to evaluate, meta-learning enables algorithms to achieve accuracy more quickly with less data. We argue that for unsupervised learning, the meta approach offers a principled means of defining and evaluating unsupervised learning.

Our main contributions are as follows. First, we show how to adapt knowledge acquired from a repository of small datasets that come from different domains, to new unsupervised datasets. Our MUL framework removes the subjectivity due to *rules of thumb* or educated guesses, and provides an objective evaluation methodology for UL. We give algorithms for various problems including making an informed decision on the number of clusters, learning common features, choosing amongst clustering algorithms, and removing outliers in a principled way.

Next, the debate on what characterizes right clustering has not made much headway since Kleinberg's impossibility framework (Kleinberg, 2003) mainly owing to the subjectivity or "artifacts of specific formalisms" (Zadeh & Ben-David, 2009; Ackerman & Ben-David, 2008). We add a fresh perspective to this debate by designing a meta-clustering algorithm that circumvents the effect of these artifacts by learning across clustering problems, and thus makes good clustering possible.

Finally, in a completely different direction, we use deep learning to automate learning of features across problems from different domains and of very different natures. We show that these seemingly unrelated problems can be leveraged to improve average performance across UL datasets. In this way, we effectively unite many heterogeneous "small data" into sufficient "big data" to benefit from a single neural network. Moreover, recently, learning for resource constrained devices (Gupta *et al.*, 2017) has attracted considerable attention, and several compression methods have been suggested specifically for deep nets, see e.g. (Chen *et al.*, 2015; Han *et al.*, 2016; Luo *et al.*, 2017). However, these methods assume homogenous data, and therefore maintaining a compressed deep net separately for each dataset would be prohibitive due to storage constrains. Our approach, in principle, may be combined with compression methods to provide a practical solution for these devices.

## 2. Learning Preliminaries

We first define agnostic learning, in general, and then define meta-learning as a special case where the examples are problems themselves. A learning task consists of a universe $\mathcal{X}$, labels $\mathcal{Y}$ and a bounded *loss function* $\ell : \mathcal{Y} \times \mathcal{Y} \to [0, 1]$, where $\ell(y, z)$ is the loss of predicting $z$ when the true label was $y$. A learner $L : (\mathcal{X} \times \mathcal{Y})^* \to \mathcal{Y}^\mathcal{X}$ takes a training set $T = \{(x_1, y_1), \ldots (x_n, y_n)\}$ consisting of a finite number of iid samples from $\mu$ and outputs a classifier $L(T) \in \mathcal{Y}^\mathcal{X}$, where $\mathcal{Y}^\mathcal{X}$ is the set of functions from $\mathcal{X}$ to $\mathcal{Y}$. The loss of a classifier $c \in \mathcal{Y}^\mathcal{X}$ is $\ell_\mu(c) = \mathrm{E}_{(x,y)\sim\mu}[\ell(y, c(x))]$, and the expected loss of $L$ is $\ell_\mu(L) = \mathrm{E}_{T\sim\mu^n}[\ell_\mu(L(T))]$. Learning is with respect to a *concept class* $\mathcal{C} \subseteq \mathcal{Y}^\mathcal{X}$.

**Definition 1** (Agnostic learning of $\mathcal{C}$). *For countable sets[1] $\mathcal{X}, \mathcal{Y}$ and $\ell : \mathcal{Y} \times \mathcal{Y} \to [0, 1]$, learner $L$ agnostically learns $\mathcal{C} \subseteq \mathcal{Y}^\mathcal{X}$ if there exists a polynomial $p$ such that for any distribution $\mu$ over $\mathcal{X} \times \mathcal{Y}$ and for any $n \geq p(1/\epsilon, 1/\delta)$, $\mathrm{Pr}_{T\sim\mu^n}[\ell_\mu(L(T)) \leq \min_{c\in\mathcal{C}} \ell_\mu(c) + \epsilon] \geq 1 - \delta$. Further, $L$ and the classifier $L(T)$ must run in time polynomial in the length of their inputs.*

PAC learning refers to the special case when $\mu$ is additional assumed to satisfy $\min_{c\in\mathcal{C}} \ell_\mu(c) = 0$.

MUL, which is the focus of this paper, simply refers to case where $\mu$ is a *meta-distribution* over datasets $X \in \mathcal{X}$ and ground truth labelings $Y \in \mathcal{Y}$. We use capital letters to represent datasets as opposed to individual examples. A meta-classifier $c$ is a UL algorithm that take an entire dataset $X$ as input, such as clustering algorithms, and outputs $Z \in \mathcal{Y}$. As mentioned, true labels need only be observed for the training datasets – we may never observe the true labels of any problem encountered after deployment. This differs from, say, online learning, where it is assumed that for each example, after you make a prediction, you find out the ground truth.

---

[1] For simplicity of presentation, we assume that these sets are countable, but with appropriate measure theoretic assumptions the analysis in this paper can be extended to the infinite case.

For a finite set $S$, $\Pi(S)$ denotes the set of clusterings or disjoint partitions of $S$ into two or more sets, e.g., $\Pi(\{1,2,3\})$ includes $\{\{1\},\{2,3\}\}$ which is the partition into clusters $\{1\}$ and $\{2,3\}$. For a clustering $C$, denote by $\cup C = \cup_{S \in C} S$ the set of points clustered. Given two clusterings $Y, Z \in \Pi(S)$, the *Rand Index* measures the fraction of pairs of points on which they agree:

$$\mathrm{RI}(Y,Z) = \frac{|\{a \neq b \in S \mid d_Y(a,b) = d_Z(a,b)\}|}{|S|(|S|-1)},$$

where for clustering $C$, we define the distance function $d_C(a,b)$ to be 0 if they are in the same cluster, i.e., $a,b \in R$ for some $R \in C$, and 1 otherwise. In our experiments, the loss we will measure is the standard *Adjusted Rand Index* (ARI) which attempts to correct the Rand Index by accounting for chance agreement (Hubert & Arabie, 1985). We denote by $ARI(Y,Z)$ the adjusted rand index between two clusterings $Y$ and $Z$. We abuse notation and also write $ARI(Y,Z)$ when $Y$ is a vector of class labels, by converting it to a clustering with one cluster for each class label. We define the loss to be the fraction of pairs of points on which they disagree, assuming the clusterings are on the same sets of points. If, for any reason the clusterings are not on the same sets of points, the loss is defined to be 1.

$$\ell(Y,Z) = \begin{cases} 1 - \mathrm{RI}(Y,Z) & if \cup Y = \cup Z \\ 1 & otherwise. \end{cases} \quad (1)$$

In *Euclidean clustering*, the points are Euclidean, so each dataset $X \subset \mathbb{R}^d$ for some $d \geq 1$. In meta-Euclidean-clustering, one aims to learn a clustering algorithm from several different training clustering problems (of potentially different dimensionalities $d$).

Rand Index measures clustering quality with respect to an *extrinsic* ground truth. In many cases, such a ground truth is unavailable, and an *intrinsic* metric is useful. Such is the case when choosing the number of clusters. Given different clusterings of size $k = 2,3,\ldots$, how can one compare and select? One popular approach is the so-called *silhouette score* (Rousseeuw, 1987), defined as follows for a Euclidean clustering:

$$\mathrm{sil}(C) = \frac{1}{|\cup C|} \sum_{x \in \cup C} \frac{b(x) - a(x)}{\max\{a(x), b(x)\}}, \quad (2)$$

where $a(x)$ denotes the average distance between point $x$ and other points in its own cluster and $b(x)$ denotes the average distance between $x$ and points in an alternative cluster, where the alternative cluster is the one (not containing $x$) with smallest average distance to $x$.

## 3. Meta-unsupervised problems

The simplest approach to MUL is Empirical Risk Minimization (ERM), namely choosing the unsupervised algorithm from some family $\mathcal{C}$ with lowest empirical error on training set $T$, which we write as $\mathrm{ERM}_\mathcal{C}(T)$. The following lemma implies a logarithmic dependence on $|\mathcal{C}|$ and helps us solve several interesting MUL problems.

**Lemma 1.** *For any finite family $\mathcal{C}$ of unsupervised learning algorithms, any distribution $\mu$ over problems $X, Y \in \mathcal{X} \times \mathcal{Y}$, and any $n \geq 1, \delta > 0$,*

$$\Pr_{T \sim \mu^n}\left[\ell_\mu(\mathrm{ERM}_\mathcal{C}(T)) \leq \min_{c \in \mathcal{C}} \ell_\mu(c) + \sqrt{\frac{2}{n}\log\frac{|\mathcal{C}|}{\delta}}\right] \geq 1 - \delta,$$

*where $\mathrm{ERM}_\mathcal{C}(T) \in \arg\min_{c \in \mathcal{C}} \sum_T \ell(Y, c(X))$ is any empirical loss minimizer over $c \in \mathcal{C}$.*

*Proof.* Fix $U_0 \in \arg\min_{U \in \mathcal{U}} \ell_\mu(U)$. Let

$$\epsilon = 2\sqrt{\frac{\log(1/\delta) + \log|\mathcal{U}|}{2m}},$$

and define $S = \{U \in \mathcal{U} \mid \ell_\mu(U) \geq \ell_\mu(U_0) + \epsilon\}$. Then Chernoff bounds imply that

$$\Pr_T\left[\frac{1}{m} \sum_{(X,Y) \in T} \ell(Y, U(X)) \geq \ell_\mu(U) + \epsilon/2\right] \leq e^{-2m(\epsilon/2)^2}.$$

Similarly, for each $U \in S$,

$$\Pr_T\left[\frac{1}{m} \sum_{(X,Y) \in T} \ell(Y, U(X)) \leq \ell_\mu(U) - \epsilon/2\right] \leq e^{-2m(\epsilon/2)^2}.$$

In order for $\ell(\mathrm{ERM}_U) \geq \min \ell(U) + \epsilon$, either some $U \in S$ must have empirical error $\leq \ell(U) - \epsilon/2$ or the empirical error of $U_0$ must be $\geq \ell(U_0) + \epsilon/2$. By the union bound, this happens with probability at most $|\mathcal{U}|e^{-2m(\epsilon/2)^2} = \delta$. □

### 3.1. Selecting the clustering algorithm/number of clusters

Instead of the ad hoc parameter selection heuristics currently used in UL, MUL provides a data-driven alternative. Suppose one has $m$ candidate clustering algorithms and/or parameter settings $C_1(X),\ldots,C_m(X)$ for each data set $X$. These may be derived from $m$ different clustering algorithms, or alternatively, they could represent the *meta-k* problem, i.e., how many clusters to choose from a single clustering algorithm where parameter $k \in \{2,\ldots,m+1\}$ determines the number of clusters. In this section, we show that choosing the right algorithm is essentially a multi-class classification problem given any set of problem metadata features and cluster-specific features. Trivially, Lemma 1 implies that with $O(\log m)$ training problem sets one can select the $C_j$ that performs best across problems. For meta-k, however, this would mean choosing the same number of clusters to use across all problems, analogous to choosing the best single class for multi-class classification.

To learn to choose the best $C_j$ on a problem-by-problem basis, suppose we have *problem features* $\phi(X) \in \Phi$ such as number of dimensions, number of points, domain (text/vision/etc.), and *cluster features* $\gamma(C_j(X)) \in \Gamma$ which might include number of clusters, mean distance to cluster center, and silhouette score (eq. 2). Suppose we also have a family $\mathcal{F}$ of functions $f : \Phi \times \Gamma^m \to \{1, 2, \ldots, m\}$, that selects the clustering based on features (i.e., any multi-class classification family may be used):

$$\arg\min_{f \in \mathcal{F}} \sum_i \ell\left(Y_i, C_{f(\phi(X_i), \gamma(C_1(X_i)), \ldots, \gamma(C_m(X_i)))}(X_i)\right).$$

The above $\text{ERM}_\mathcal{F}$ is simply a reduction from the problem of selecting $C_j$ from $X$ to the problem of multi-class classification based on features $\phi(X)$ and $\gamma(C_1(X)), \ldots, \gamma(C_m(X))$ and loss as defined in eq. (1). As long as $\mathcal{F}$ can be parametrized by a fixed number of $b$-bit numbers, the ERM approach of choosing the "best" $f$ will be statistically efficient. If $\text{ERM}_\mathcal{F}$ cannot be computed exactly within time constraints, an approximate minimizer may be used.

**Fitting the threshold in single-linkage clustering.** To illustrate a concrete efficient algorithm, consider choosing the threshold parameter of a single linkage clustering algorithm. Fix the set of possible vertices $\mathcal{V}$. Take $\mathcal{X}$ to consist of undirected weighted graphs $X = (V, E, W)$ with vertices $V \subseteq \mathcal{V}$, edges $E \subseteq \{\{u, v\} \mid u, v \in V\}$ and non-negative weights $W : E \to \mathbb{R}_+$. The loss on clusterings $\mathcal{Y} = \Pi(\mathcal{V})$ is again as defined in Eq. (1). Note that Euclidean data could be transformed into, e.g., the complete graph with $W(\{x, x'\}) = \|x - x'\|$.

Single-linkage clustering with parameter $r \geq 0$, $C_r(V, E, w)$ partitions the data into connected components of the subgraph of $(V, E)$ consisting of all edges whose weights are less than or equal to $r$. For generalization bounds, we simply assume that numbers are represented with a constant number of bits, as is common today.

**Theorem 1.** *The class $\{C_r \mid r > 0\}$ of single-linkage algorithms with threshold $r$ where numbers are represented using $b$ bits, can be agnostically learned. In particular, a quasilinear time algorithm achieves error $\leq \min_r \ell_\mu(C_r) + \sqrt{2(b + \log 1/\delta)/n}$, with prob. $\geq 1 - \delta$ over $n$ training problems.*

*Proof.* For generalization, we assume that numbers are represented using at most $b$ bits. By Lemma 1, we see that with $m$ training graphs and $|\{C_r\}| \leq 2^b$, we have that with probability $\geq 1 - \delta$, the error of ERM is within $\sqrt{2(b + \log 1/\delta)/n}$ of $\min_r \ell_\mu(C_r)$.

It remains to show how one can find the best single-linkage parameter in quasilinear time. It is trivial to see that one can find the best cutoff for $r$ in polynomial time: for each edge weight $r$ in the set of edge weights across all graphs, compute the mean loss of $C_r$ across the training set. Since $C_r$ runs in polynomial time, loss can be computed in polynomial time, and the number of different possible cutoffs is bounded by the number of edge weights which is polynomial in the input size, the entire algorithm runs in polynomial time.

For a quasilinear-time algorithm (in the input size $|T| = \Theta(\sum_i |V_i|^2)$), run Kruskal's algorithm on the union graph of all of the graphs in the training set (i.e., the number of nodes and edges are the sum of the number of nodes and edges in the training graphs, respectively). As Kruskal's algorithm adds each new edge to its forest (in order of non-decreasing edge weight), effectively two clusters in some training graph $(V_i, E_i, W_i)$ have been merged. The change in loss of the resulting clustering can be computed from the loss of the previous clustering in time proportional to the product of the two clusters that are being merged, since these are the only values on which $Z_i$ changed. Naïvely, this may seem to take order $\sum_i |V_i|^3$. However, note that, each pair of nodes begins separately and is updated, exactly once during the course of the algorithm, to be in the same cluster. Hence, the total number of updates is $O(\sum_i |V_i|^2)$, and since Kruskal's algorithm is quasilinear time itself, the entire algorithm is quasilinear. For correctness, it is easy to see that as Kruskal's algorithm runs, $C_r$ has been computed for each possible $r$ at the step just preceding when Kruskal adds the first edge whose weight is greater than $r$. □

### 3.2. Outlier removal

For simplicity, we consider learning the single hyperparameter of the fraction of examples, furthest from the mean, to remove. In particular, suppose training problems are classification instances, i.e., $X_i \in \mathbb{R}^{d_i \times m_i}$ and $Y_i \in \{1, 2, \ldots, k_i\}^{m_i}$. To be concrete, suppose one is using algorithm $C$ which is, say, K-means clustering. Choosing the parameter $\theta$ which is the fraction of outliers to ignore during fitting, one might define $C_\theta$ with parameter $\theta \in [0, 1)$ on data $x_1, \ldots, x_n \in \mathbb{R}^d$ as follows: (a) compute the data mean $\mu = \frac{1}{n} \sum_i x_i$, (b) set aside as outliers the $\theta$ fraction of examples where $x_i$ is furthest from $\mu$ in Euclidean distance, (c) cluster the data with outliers removed using $C$, and (d) assign each outlier to the nearest cluster center.

We can trivially choose the best $\theta$ so as to optimize performance. With a single $b$-bit parameter $\theta$, Lemma 1 implies that this choice of $\theta$ will give a loss within $\sqrt{2(b + \log 1/\delta)/n}$ of the optimal $\theta$, with probability $\geq$ $1 - \delta$ over the sample of datasets. The number of $\theta$'s that need to be considered is at most the total number of inputs across problems, so the algorithm runs in polynomial time.

### 3.3. Problem recycling

For this model, suppose that each problem belongs to a set of common problem categories, e.g., digit recognition, sentiment analysis, image classification among the thousands of classes of ImageNet (Russakovsky *et al.*, 2015), etc. The idea is that one can recycle the solution to one version of the problem in a later incarnation. For instance, suppose that one trained a digit recognizer on a previous problem. For a new problem, the input may be encoded differently (e.g., different image size, different pixel ordering, different color representation), but there is a transformation $T$ that maps this problem into the same latent space as the previous problem so that the prior solution can be re-used. In particular, for each problem category $i = 1, 2, \ldots, N$, there is a latent problem space $\Lambda_i$ and a solver $S_i : \Lambda_i \to \mathcal{Y}_i$. Each problem $X, Y$ of this category can be transformed to $T(X) \in \Lambda_i$ with low solution loss $\ell(Y, S(T(X)))$. In addition to the solvers, one also requires a *meta-classifier* $M : \mathcal{X} \to \{1, 2, \ldots, N\}$ that, for a problem $X$, identifies which solver $i = M(X)$ to use. Finally, one has transformers $T_i : M^{-1}(i) \to L_i$ that map any $X$ such that $M(X) = i$ into latent space $\Lambda_i$. The output of the meta-classifier is simply $S_{M(X)}(T_{M(X)}(X))$. Lemma 1 implies that if one can optimize over meta-classifiers and the parameters of the meta-classifier are represented by $D$ $b$-bit numbers, then one achieves loss within $\epsilon$ of the best meta-classifier with $m = O\left(Db/\epsilon^2\right)$ problems.

## 4. The possibility of meta-clustering

In this section, we point out how the framing of meta-clustering circumvents Kleinberg's impossibility theorem for clustering. To review, (Kleinberg, 2003) considers clustering finite sets of points $X$ endowed with symmetric distance functions $d \in D(X)$, where the set of valid distance functions is:

$$D(X) = \{d : X \times X \to \mathbb{R} \mid \forall x, x' \in X \\ d(x, x') = d(x', x) \geq 0, d(x, x') = 0 \text{ iff } x = x'\}. \quad (3)$$

A clustering algorithm $A$ takes a distance function $d \in D(X)$ and returns a partition, i.e., $A(d) \in \Pi(X)$. Kleinberg defines an axiomatic framework with the following three desirable properties, and proves that no clustering algorithm $A$ can satisfy all these properties simultaneously:

**Scale-Invariance**. For any distance function $d$ and any $\alpha > 0$, $A(d) = A(\alpha \cdot d)$, where $\alpha \cdot d$ is the distance function $d$ scaled by $\alpha$. That is, the clustering should not change if the problem is scaled by a constant factor.

**Richness**. For any finite $X$ and clustering $C \in \Pi(X)$, there exists $d \in D(X)$ such that $A(d) = C$. Richness implies that for any partition there is an arrangement of points where that partition is the correct clustering.

**Consistency**. Let $d, d' \in D(X)$ such that $A(d) = C$, and for all $x, x' \in X$, if $x, x'$ are in the same cluster in $C$ then $d'(x, x') \leq d(x, x')$ while if $x, x'$ are in different clusters in $C$ then $d'(x, x') \geq d(x, x')$. Then the axiom demands $A(d') = A(d)$. That is, the clustering should not change if the points within a cluster are pulled closer to each other, and the points in different clusters are pushed further apart.

For intuition, consider clustering two points where there is a single distance. Should they be in a single cluster or two clusters? By richness, there must be some distances $\delta_1, \delta_2 > 0$ such that if $d(x_1, x_2) = \delta_1$ then they are in the same cluster while if $d(x_1, x_2) = \delta_2$ they are in different clusters. This, however violates Scale-Invariance, since the two problems are at a scale $\alpha = \delta_2/\delta_1$ of each other. We show that a natural meta-version of the axioms is satisfied by a simple meta-single-linkage clustering algorithm. The main insight is that prior problems can be used to define a scale in the meta-clustering framework. Suppose we define the clustering problem with respect to a non-empty training set of clustering problems. So a meta-clustering algorithm takes $t \geq 1$ training clustering problems $M(d_1, C_1, \ldots, d_t, C_t) = A$ with their ground-truth clusterings (on corresponding sets $X_i$, i.e., $d_i \in D(X_i)$ and $C_i \in \Pi(X_i)$) and outputs a clustering algorithm $A$. We can use these training clusterings to establish a scale. In particular, we will show a meta-clustering algorithm whose output $A$ always satisfies the second two axioms and which satisfies the following variant of Scale-Invariance:

**Meta-Scale-Invariance**. Fix any distance functions $d_1, d_2, \ldots, d_t$ and ground truth clusterings $C_1, \ldots, C_t$ on sets $X_1, \ldots, X_t$. For any $\alpha > 0$, and any distance function $d$, if $M(d_1, C_1, \ldots, d_t, C_t) = A$ and $M(\alpha \cdot d_1, C_1, \ldots, \alpha \cdot d_t, C_t) = A'$, then $A(d) = A'(\alpha \cdot d)$.

**Theorem 2.** *There is a meta-clustering algorithm that satisfies Meta-Scale-Invariance and whose output always satisfies Richness and Consistency.*

*Proof.* There are a number of such clustering algorithms, but for simplicity we create one based on single-linkage clustering. Single-linkage clustering satisfies Richness and Consistency (see (Kleinberg, 2003), Theorem 2.2). With meta-clustering, the scale can be established using training data. The question is how to choose its single-linkage parameter. One can choose it to be the minimum distance between any two points in different clusters across all train-ing problems. It is easy to see that if one scales the training problems and $d$ by the same factor $\alpha$, the clusterings remain unchanged, and hence the meta-clustering algorithm satisfies meta-scale-invariance. □

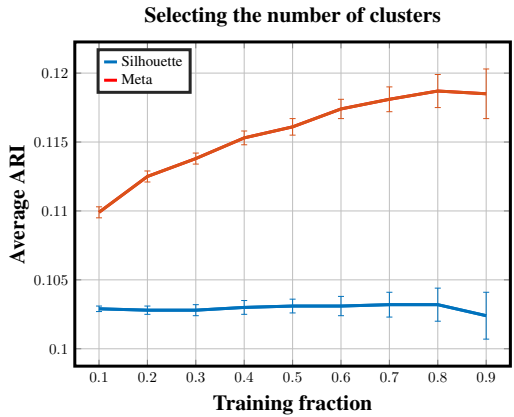
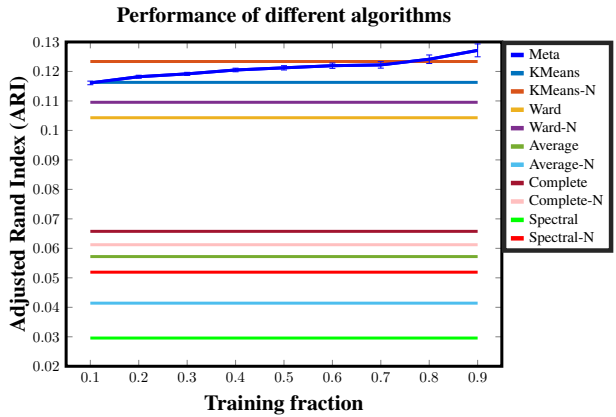

*Figure 2.* Average ARI scores of the meta-algorithm and the baseline for choosing the number of clusters, versus the fraction of problems used for training. We observe that the best $k$ predicted by the meta approach registered a much higher ARI than that by Silhouette score maximizer. Also, as shown in the Supplementary, the meta-algorithm achieved a much lower root-mean-square error.

*Figure 3.* ARI scores of different clustering ($k$=2) algorithms on OpenML binary classification problems. The meta algorithm (95% confidence intervals are shown) is compared with standard clustering baselines on both the original data as well as the transformed data (denoted by "-N") where each feature is normalized to have zero mean and unit variance. The meta-algorithm, given sufficient training problems, is able to outperform all the other algorithms.

## 5. Experiments

We conducted several experiments to substantiate the efficacy of the proposed framework under various unsupervised settings. We downloaded all classification datasets from OpenML (http://www.openml.org) that had at most 10,000 instances, 500 features, 10 classes, and no missing data to obtain a corpus of 339 datasets. We now describe in detail the results of our experiments, highlighting the effect of our meta paradigm in improving the performance.

### 5.1. Selecting the number of clusters

For the purposes of this section, we fix the clustering algorithm to be K-means and compare two approaches to choosing the number of clusters, $k$ from $k = 2$ to 10. More generally, one could vary $k$ on, for instance, a logarithmic scale, or a combination of different scales. First, we consider a standard heuristic for the baseline choice of $k$: for each cluster size $k$ and each dataset, we generate 10 clusterings from different random starts for K-means and take one with best Silhouette score among the 10. Then, over the 9 different values of $k$, we choose the one with greatest Silhouette score so that the resulting clustering is the one of greatest Silhouette score among all 90. Similar to the approach for choosing the best algorithm above, in our meta-k approach, the meta-algorithm outputs $\hat{k}$ as a function of Silhouette score and $k$ by outputting the $\hat{k}$ with greatest estimated ARI. We evaluate on the same 90 clusterings for the 339 datasets as the baseline. To estimate ARI in this experiment, we used simple least-squares linear regression. In particular, for each $k \in \{2, \ldots, 9\}$, we fit ARI as a linear function of Silhouette scores using all the data from the meta-training set in the partition pertaining to $k$: each dataset in the meta-training set provided 10 target values, corresponding to different runs where number of clusters was fixed to $k$. As is standard in the clustering literature, we define the best-fit $k_i^*$ for dataset $i$ to be the one that yielded maximum ARI score across the different runs, which is often different from $k_i$, the number of clusters in the ground truth (i.e., the number of class labels). We held out a fraction of the problems for test and used the remaining for training. We evaluated two quantities of interest: the ARI and the root-mean-square error (RMSE) between $\hat{k}$ and $k^*$. Both quantities were better for the meta-algorithm than the baseline where we performed 1000 train-test splits to compute the confidence intervals (Fig. 2).

### 5.2. Selecting the clustering algorithm

We now consider the question of which of a number of given clustering algorithms to use to cluster a given data set. We illustrate the main ideas with $k = 2$ clusters. First, one can run each of the algorithms on the repository and see which algorithm has the lowest average error. Error is calculated with respect to the ground truth labels by the ARI (see Section 2). We compare algorithms on the 250 openml binary classification datasets with at most 2000 instances. The ten base clustering algorithms were chosen to be five clustering algorithms from scikit-learn (Pedregosa *et al.*, 2011): K-Means, Spectral, Agglomerative Single Linkage, Complete Linkage, and Ward, together with a second version of each in which each attribute is first normalized to have zero mean and unit variance. Each algorithm is run with the default scikit-learn parameters. We implement the algorithm selection approach of Section 3, learning to choose a different algorithm for each problem based on problem and cluster-specific features. Given clustering $\Pi$ of $X \in \mathbb{R}^{d \times m}$, the feature vector $\Phi(X, \Pi)$ consists of the dimensionality, number of examples, minimum and maximum eigenvalues of the covariance matrix, and the silhouette score (see Section 2) of the clustering $\Pi$:

$$\Phi(X, \Pi) = (d, m, \sigma_{\min}(\Sigma(X)), \sigma_{\max}(\Sigma(X)), \text{sil}(\Pi)),$$

where $\Sigma(X)$ denotes the covariance matrix of $X$, and $\sigma_{\min}(M)$ and $\sigma_{\max}(M)$ denote the minimum and maximum eigenvalues, respectively, of matrix $M$. Instead of choosing the clustering with best Silhouette score, which is a standard approach, the meta-clustering algorithm effectively learns terms that can correct for over- or under-estimates, e.g., learning for which problems the Silhouette heuristic tends to produce too many clusters. To choose which of the ten clustering algorithms on each problem, we fit ten estimators of accuracy (ARI, see Section 2) based on these features. That is for each clustering algorithm $C_j$, we fit $\text{ARI}(Y_i, C_j(X_i))$ from features $\Phi(X_i, C_j(X_i)) \in \mathbb{R}^5$ over problems $X_i, Y_i$ using $\nu$-SVR regression, with default parameters as implemented by scikit-learn. Call this estimator $\hat{a}_j(X, C_j(X))$. To cluster a new dataset $X \in \mathbb{R}^{d \times m}$, the meta-algorithm then chooses $C_j(X)$ for the $j$ with greatest accuracy estimate $\hat{a}_j(X, C_j(X))$. The 250 problems were divided into train and test sets of varying sizes. The results, shown in Figure 3, demonstrate two interesting features. First, one can see that the different baseline clustering algorithms had very different average performances, suggesting that a principled approach like ours to select algorithms can make a difference. Further, Figure 3 shows that the meta-algorithm, given sufficiently many training problems, is able to outperform, on average, all the baseline algorithms. This is despite the fact that the 250 problems have different dimensionalities and come from different domains. Based on the trend, it is possible that with more training problems it would further surpass the baselines.

### 5.3. Removing outliers

We also experimented to see if removing outliers improved average performance on the same 339 classification problems. Our objective was to choose a single best fraction to remove from all the meta-test sets. For each data set $X$, we removed a $p \in \{0, 0.01, 0.02, \ldots, 0.05\}$ fraction examples with the highest euclidean norm in $X$ as outliers, and like-wise for each meta-test set in the partition. The previous section corresponds exactly to the case $p = 0$. We first clustered the data without outliers, and obtained the corresponding Silhouette scores. We then put back the outliers by assigning them to their nearest cluster center, and computed the ARI score thereof. Then, following an identical procedure to the meta-k algorithm of Section 5.1, we fitted regression models for ARI corresponding to complete data using the silhouette scores on pruned data, and measured the effect of outlier removal in terms of the true average ARI (corresponding to the best predicted ARI) over entire data. Again, we report the results averaged over 10 independent train/test partitions. As Fig. 4 shows, by treating 1% of the instances in each dataset as outliers, we achieved an improvement in ARI scores relative to clustering with all the data as in section 5.1. As the fraction of data considered outlier was increased to 2% or higher, however, performance degraded. Clearly, we can learn what fraction to remove based on data, and improve the performance considerably even with such a simple algorithm.

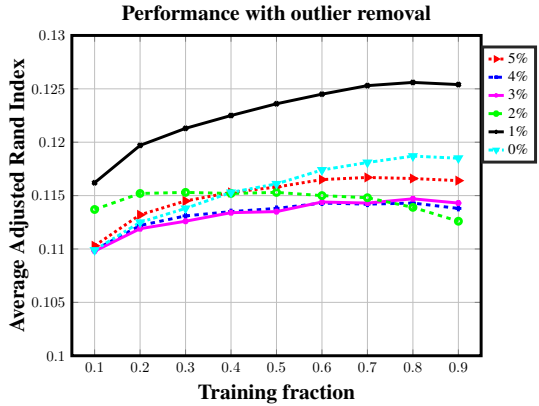

*Figure 4.* Outlier removal results. Removing 1% of the instances as outliers improves on the ARI scores obtained without outlier removal. Interestingly, going beyond 1% decreases the performance to that without outlier removal. Such data dependent observations would be missed by the prevalent rules of thumb. Our algorithm naturally figures out the right proportion (1%) in a principled way.

### 5.4. Deep learning binary similarity function

In this section, we consider a new unsupervised problem of learning a binary similarity function (BSF) that predicts whether two examples from a given problem should belong to the same cluster (i.e., have the same class label). Formally, a problem is specified by a set $X$ of data and meta-features $\phi$. The goal is to learn a classifier $f(x, x', \phi) \in \{0, 1\}$ that takes two examples $x, x' \in X$ and the corresponding problem meta-features $\phi$, and predicts 1 if the input pair would belong to the same cluster (or have the same class la-bels). In our experiments, we take Euclidean data $X \subseteq \mathbb{R}^d$ (each problem may have different dimensionality $d$), and the meta-features $\phi = \Sigma(X)$ consist of the covariance matrix of the unlabeled data. We restricted our experiments to the 146 datasets with at most 1000 examples and 10 features. We normalized each dataset to have zero mean and unit variance along every coordinate (hence, every diagonal element in the covariance matrix was set to 1). We randomly sampled pairs of examples from each dataset to form disjoint *meta-training* and *meta-test* sets. For each pair, we concatenated the features to create data with 20 features (padding examples with fewer than 10 features with zeros). We then computed the empirical covariance matrix of the dataset, and vectorized the entries of the covariance matrix on and above the leading diagonal to obtain an additional 55 covariance features. Concatenating these features with the 20 features formed a 75-dimensional feature vector per pair. Thus all pairs sampled from the same dataset shared the 55 covariance features. Moreover, we derived a new binary label dataset in the following way. We assigned a label 1 to pairs formed by combining examples belonging to the same class, and 0 to those resulting from the different classes.

Each dataset in the first category was used to sample data pairs for both the meta-training and the meta-internal_test (meta-IT) datasets, while the second category did not contribute any training data and was exclusively used to generate only the meta-external_test (meta-ET) dataset. Our procedure ensured a disjoint intersection between the meta-training and the meta-IT data, and resulted in 10 separate (meta-training, meta-IT, meta-ET) triplets. Note that combining thousands of examples from each of hundreds of problems yields hundreds of thousands of examples, **turning small data into big data**. This provides a means of making DNNs naturally applicable to data sets that might have otherwise been too small. For each meta-training set, we trained an independent deep net model. The complete details of the sampling process, the network architecture, and the training procedure are given in the Supplementary.

We tested our trained models on meta-IT and meta-ET data. For each feature vector in meta-IT (respectively meta-ET), we computed the predicted same class probability. We added the predicted same class probability for the feature vector obtained with flipped order, as described earlier for the feature vectors in the meta-training set. We predicted the instances in the corresponding pair to be in the same cluster only if the average of these two probabilities exceeded 0.5. We compared the meta approach to a hypothetical majority rule that had prescience about the class distribution. As the name suggests, the majority rule predicted all pairs to have the majority label, i.e., on a problem-by-problem basis we determined whether 1 (same class) or 0 (different class) was more accurate and gave the baseline the advantage of this knowledge for each problem, even though it normally would not be available at classification time. This information about the distribution of the labels was not accessible to our meta-algorithm. Fig. 5 shows the average fraction of similarity pairs correctly identified relative to the corresponding pairwise ground truth relations on the two test sets, and the corresponding standard deviations across the 10 independent (meta-training, meta-IT, meta-ET) collections. Clearly, the meta approach outperforms the majority rule on meta-IT, illustrating the benefits of the meta approach in a multi-task transductive setting. More interesting, still, is the significant improvement exhibited by the meta method on meta-ET, despite having its category precluded from contributing any data for training. The result clearly demonstrates the benefits of leveraging archived supervised data for informed decision making in unsupervised settings.

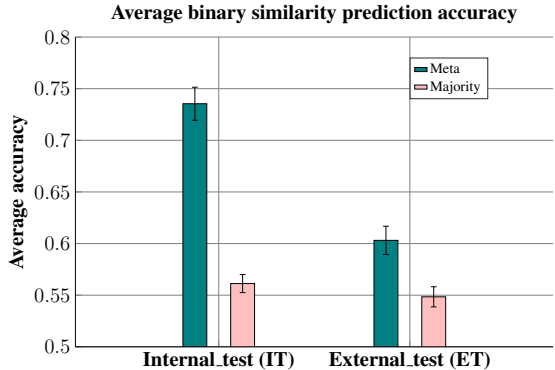

*Figure 5.* Mean accuracy and standard deviation on meta-IT and meta-ET data. Comparison between the fraction of pairs correctly predicted by the meta algorithm and the majority rule. Recall that meta-ET, unlike meta-IT, was generated from a partition that did not contribute any training data. Nonetheless, the meta approach significantly improved upon the majority rule even on meta-ET.

## 6. Conclusion

Treating each UL problem in isolation has made UL notoriously difficult to define. We suggest that UL problems be viewed as representative samples from some meta-distribution $\mu$. We show how a repository of multiple datasets annotated with ground truth labels can be used to improve average performance on several unsupervised tasks, even with simple algorithms. Theoretically, this enables us to make UL problems, such as clustering, well-defined.

Prior datasets may prove useful for a variety of reasons, from simple to complex. They may help one choose the best clustering algorithm or parameter settings, or transfer shared features that can be identified as useful. We demonstrate how the combination of many small heterogeneous data sets can form a large data set appropriate for training with a common deep network. This lets us accomplish zero shot learning across *small and diverse* data in a natural way.

# A. Complete details: Deep learning a binary similarity function

We consider a new unsupervised problem of learning a binary similarity function (BSF) that predicts whether two examples from a given problem should belong to the same cluster (i.e., have the same class label). Formally, a problem is specified by a set $X$ of data and meta-features $\phi$. The goal is to learn a classifier $f(x, x', \phi) \in \{0, 1\}$ that takes two examples $x, x' \in X$ and the corresponding problem meta-features $\phi$, and predicts 1 if the input pair would belong to the same cluster (or have the same class labels).

In our experiments, we take Euclidean data $X \subseteq \mathbb{R}^d$ (each problem may have different dimensionality $d$), and the meta-features $\phi = \Sigma(X)$ consist of the covariance matrix of the unlabeled data. We restricted our experiments to the 146 datasets with at most 1000 examples and 10 features. We normalized each dataset to have zero mean and unit variance along every coordinate (hence, every diagonal element in the covariance matrix was set to 1).

We randomly sampled pairs of examples from each dataset to form *meta-training* and *meta-test* sets as described in the following section. For each pair, we concatenated the features to create data with 20 features (padding examples with fewer than 10 features with zeros). We then computed the empirical covariance matrix of the dataset, and vectorized the entries of the covariance matrix on and above the leading diagonal to obtain an additional 55 covariance features. Concatenating these features with the 20 features formed a 75-dimensional feature vector per pair. Thus all pairs sampled from the same dataset shared the 55 covariance features. Moreover, we derived a new binary label dataset in the following way. We assigned a label 1 to pairs formed by combining examples belonging to the same class, and 0 otherwise.

### A.0.1. SAMPLING PAIRS TO FORM META-TRAIN AND META-TEST DATASETS

We formed a partition of the new datasets by randomly assigning each dataset to one of the two categories with equal probability. Each dataset in the first category was used to sample data pairs for both the meta-training and the meta-internal_test (meta-IT) datasets, while the second category did not contribute any training data and was exclusively used to generate only the meta-external_test (meta-ET) dataset.

We constructed meta-training pairs by sampling randomly pairs from each dataset in the first category. In order to mitigate the bias resulting from the variability in size of the different datasets, we restricted the number of pairs sampled from each dataset to at most 2500. Likewise, we obtained the meta-IT dataset by collecting randomly sampling each dataset subject to the maximum 2500 pairs. Specifically, we randomly shuffled each dataset belonging to the first category, and used the first half (or 2500 examples, whichever was fewer) of the dataset for the meta-training data, and the following indices for the meta-IT data, again subject to maximum 2500 instances. This procedure ensured a disjoint intersection between the meta-training and the meta-IT data. Note that combining thousands of examples from each of hundreds of problems yields hundreds of thousands of examples, thereby turning small data into big data. This provides a means of making DNNs naturally applicable to data sets that might have otherwise been too small.

We created the meta-ET data using datasets belonging to the second category. Again, we sampled at most 2500 examples from each dataset in the second category. We emphasize that the datasets in the second category did not contribute any training data for our experiments.

We performed 10 independent experiments to obtain multiple partitions of the datasets into two categories, and repeated the aforementioned procedure to prepare 10 separate (meta-training, meta-IT, meta-ET) triplets. This resulted in the following (average size +/- standard deviation) statistics for dataset sizes:

$$\text{meta-training and meta-IT} : 1.73 \times 10^5 \pm 1.07 \times 10^4$$
$$\text{meta-ET} : 1.73 \times 10^5 \pm 1.19 \times 10^4$$

In order to ensure symmetry of the binary similarity function, we introduced an additional meta-training pair for each meta-training pair in the meta-training set: in this new pair, we swapped the order of the feature vectors of the instances while replicating the covariance features of the underlying dataset that contributed the two instances (note that since the covariance features were symmetric, they carried over unchanged).

### A.0.2. TRAINING NEURAL MODELS

For each meta-training set, we trained an independent deep net model with 4 hidden layers having 100, 50, 25, and 12 neurons respectively over just 10 epochs, and used batches of size 250 each. We updated the parameters of the model via the Adadelta (Zeiler, 2012) implementation of the stochastic gradient descent (SGD) procedure supplied with the Torch library[2] with the default setting of the parameters, specifically, interpolation parameter equal to 0.9 and no weight decay. We trained the model via the standard negative log-likelihood criterion (NLL). We employed ReLU non-linearity at each hidden layer but the last one, where we invoked the log-softmax function.

We tested our trained models on meta-IT and meta-ET data. For each feature vector in meta-IT (respectively meta-ET), we computed the predicted same class probability. We added the predicted same class probability for the feature vector obtained with flipped order, as described earlier for the feature vectors in the meta-training set. We predicted the instances in the corresponding pair to be in the same cluster if the average of these two probabilities exceeded 0.5, otherwise we segregated them.

---
[2]See https://github.com/torch/torch7.

### A.0.3. RESULTS

We compared the meta approach to a hypothetical majority rule that had prescience about the class distribution. As the name suggests, the majority rule predicted all pairs to have the majority label, i.e., on a problem-by-problem basis we determined whether 1 (same class) or 0 (different class) was more accurate and gave the baseline the advantage of this knowledge for each problem, even though it normally wouldn't be available at classification time. This information about the distribution of the labels was not accessible to our meta-algorithm.

Fig. 5 shows the average fraction of similarity pairs correctly identified relative to the corresponding pairwise ground truth relations on the two test sets, and the corresponding standard deviations across the 10 independent (meta-training, meta-IT, meta-ET) collections. Clearly, the meta approach outperforms the majority rule on meta-IT, illustrating the benefits of the meta approach in a multi-task transductive setting. More interesting, still, is the significant improvement exhibited by the meta method on meta-ET, despite having its category precluded from contributing any data for training. The result clearly demonstrates the benefits of leveraging archived supervised data for informed decision making in unsupervised settings such as binary similarity prediction.

## B. Additional results (Selecting the number of clusters)

Fig. 6 shows how the root-mean-square error (RMSE) between $\hat{k}$ and $k^*$ when we performed 1000 train-test splits to compute the confidence intervals. Clearly, the meta-algorithm registers a much lower RMSE than the Silhouette baseline. Moreover the discrepancy between the two methods increases, in favor of our algorithm, with an increase in the training data.

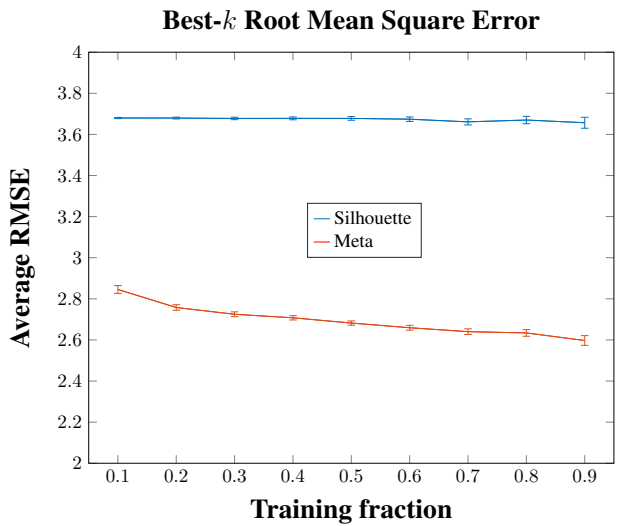

*Figure 6.* RMSE of distance to the best $k$, i.e. $k^*$, across datasets, i.e., comparing $\sqrt{\frac{1}{n}\sum(\tilde{k}_i - k_i^*)^2}$ and $\sqrt{\frac{1}{n}\sum(\hat{k}_i - k_i^*)^2}$, where $\hat{k}$ and $\tilde{k}$ are the output of the meta-k algorithm and Silhouette baseline respectively. Clearly, the meta-k method outputs a number of clusters much closer to $k^*$ than the Silhouette score maximizer.